\documentclass[conference]{IEEEtran}
\IEEEoverridecommandlockouts
\usepackage{cite}
\usepackage{pythontex}
\usepackage{amsmath,amssymb,amsfonts}
\usepackage{algorithmic}
\usepackage{graphicx}
\usepackage{textcomp}
\usepackage{xcolor}
\def\BibTeX{{\rm B\kern-.05em{\sc i\kern-.025em b}\kern-.08em
    T\kern-.1667em\lower.7ex\hbox{E}\kern-.125emX}}
\begin{document}
\makeatletter

\title{Train and Deploy an Image Classifier for Disaster Response\\
}

\makeatletter
\newcommand{\linebreakand}{%
  \end{@IEEEauthorhalign}
  \hfill\mbox{}\par
  \mbox{}\hfill\begin{@IEEEauthorhalign}
}
\makeatother

\author
{\IEEEauthorblockN{Jianyu Mao}
\IEEEauthorblockA{\textit{The Pennsylvania State University} \\
\textit{jxm6165@psu.edu}\\
\textit{0000-0002-8752-9403}\\
}
\and
\IEEEauthorblockN{Kiana Harris}
\IEEEauthorblockA{\textit{The Pennsylvania State University} \\
\textit{kyh5354@psu.edu}\\
\textit{0000-0002-8016-1351}\\
}
\and
\IEEEauthorblockN{Nae-Rong Chang}
\IEEEauthorblockA{\textit{The Pennsylvania State University} \\
\textit{nxc57@psu.edu}\\
\textit{0000-0002-0733-4586}\\
}
\linebreakand
\IEEEauthorblockN{Caleb Pennell}
\IEEEauthorblockA{\textit{The Pennsylvania State University} \\
\textit{cwp5377@psu.edu}\\
\textit{0000-0002-7898-4332}\\
}
\and
\IEEEauthorblockN{Yiming Ren}
\IEEEauthorblockA{\textit{The Pennsylvania State University} \\
\textit{yzr12@psu.edu}\\
\textit{0000-0001-9475-8581}\\
}
}

\maketitle

\begin{abstract}
With Deep Learning Image Classification becoming more powerful each year, it is apparent that its introduction to disaster response will increase the efficiency that responders can work with. Using several Neural Network Models, including AlexNet, ResNet, MobileNet, DenseNets, and 4-Layer CNN, we have classified flood disaster images from a large image data set with up to 79\% accuracy.  Our models and tutorials for working with the data set have created a foundation for others to classify other types of disasters contained in the images.
\end{abstract}


\section{Introduction}
After Hurricane Maria struck Puerto Rico, researchers from MIT’s Lincoln Laboratory were hard at work helping the Federal Emergency Management Agency, also known as FEMA, assess the damage. This is when the MIT researchers came up with a large LADI data set, also known as the Low Altitude Disaster Imagery data set \cite{bladi}.  In the initial state, LADI focused on the Atlantic hurricane and coastal states along the Atlantic Ocean and Gulf of Mexico.  However, this initial data set arose with various issues revolving around image sorting and misidentifying images from recognition systems.

In any large-scale disaster scenario, teams of emergency responders like FEMA could save significant time and resources by reviewing the conditions prior. Therefore, the project was organized into two goals that leveraged the data set. In the initial state, the data set consisted of human and machine annotated aerial images collected by the Civil Air Patrol in support of various disaster responses from 2015-2019. The first goal was to develop deep learning models for image classification using the LADI data set to prioritize flooding, debris, buildings, and other infrastructures. The second goal was to make our deep learning models available publicly to enable potential end-users to adopt, modify, and even improve our models.  

\section{Method}
\subsection{Data Processing}

The LADI data set contains more than 200,000 data points, and each image is labeled as one of the 6 categories - Damage, Rubble, Landslide, Flooding, Road Washout, and Fire. 
With our goal to create an image classification algorithm to correctly identify disaster response, a sufficient data set becomes the most valuable thing for us to construct an accurate deep learning model.

\subsubsection{Data Cleaning and Validation}
When it comes to real-world data, it is not improbable that data may contain incomplete, inconsistent, or missing values. If the data is corrupted, then the model might fail to yield ideal results.  
To create a reliable data set, our main aim of data cleaning is to identify and remove errors and duplicate data. This will improve our data quality and enable accurate decision making. 

Besides data cleaning, we have also restricted the LADI data set with only flooding and non-flooding images. For this step, we only focus on the metadata and label files. After generating the data set with only flooding information, we have extracted the data set into 2000 images and stored them into a separate file for model implementation.

Steps for data cleaning and validation:
\begin{itemize}
\item Extract labels with damage and infrastructure categories
\item Filter out infrastructure label with the label 'none'
\item Extract data with the label that contains 'flood'
\item Extract S3 URL data with the label that contains 'flood'
\item Extract URL data with the label that does not contain 'flood'
\end{itemize}

\subsubsection{Data Augmentation}

Having a large data set is always beneficial for the performance of the deep learning model. By utilizing the transform functions in the TorchVision package\cite{torch}, it can help to increase the amount of relevant data in our data set. 

We have used the following transform functions: 
\begin{itemize}
\item transforms.Resize(256): Resize the input image with width to be 256 pixels.
\item transforms.RandomRotation(10): Rotate the input image by a random angle not greater than 10 $^{\circ}$.
\item transforms.RandomCrop(250): Randomly crop the images to the size of $250\times250$ pixels.
\item transforms.RandomHorizontalFlip(): Horizontally flip the given PIL Image randomly with a given probability (50\% if no parameter specified).
\end{itemize}

\begin{figure}[htbp]
\centerline{\includegraphics[scale=0.35]{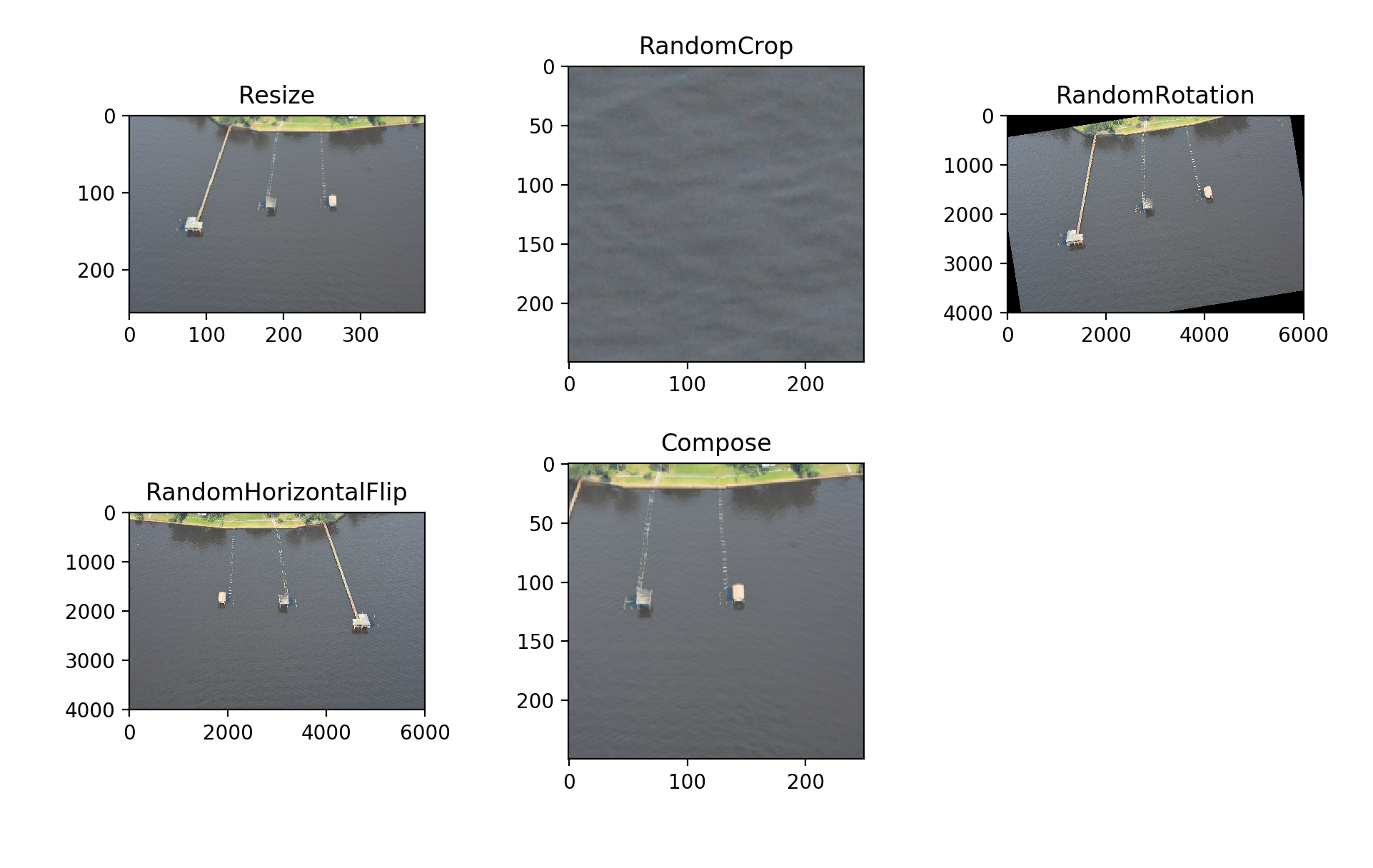}}
\caption{Five commonly used TorchVision parameter for image augmentation}
\label{fig:transform}
\end{figure}

\subsection{Model}
Deep learning models consist of diverse neural network architectures. Among them, Convolutional Neural Networks (CNNs) are most commonly used to analyze visual imagery and perform image classification tasks. The main success of utilizing CNNs for image classification is to get a comprehensive understanding and use of digital image processing techniques. In this section, we introduce and illustrate the regular CNN architecture and advanced networks, such as AlexNet, ResNet, DenseNet and MobileNet, which will be used in our experiments discussed in the next section.

\subsubsection{Convolutional Neural Network}

A Convolutional Neural Network (CNN) \cite{bcnn1} \cite{bcnn2} is the most prevalent neural network model being used for image classification tasks. A CNN architecture consists of alternate convolutional layers and pooling layers that are followed by fully-connected layers to generate outputs. The structure of a CNN model is shown in Fig.~\ref{fig:cnn}.

\begin{figure}[htbp]
\centerline{\includegraphics[scale=0.24]{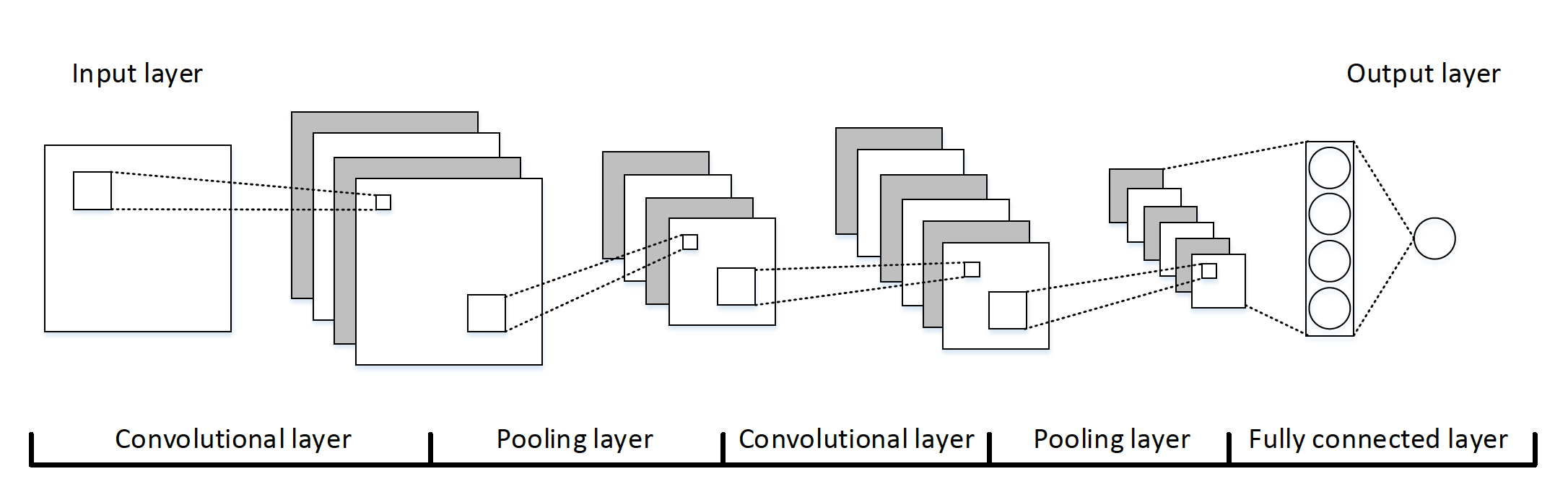}}
\caption{Structure of a Convolutional Neutral Network.}
\label{fig:cnn}
\end{figure}

\begin{itemize}
    \item Convolutional Layers: Convolutional layers convolve the input and pass the result to the next layer. The use of convolution operations is also the source for the name of this kind of architecture. Instead of using fully connected layers to learn from each pixel resulting in numerous free parameters of weights, CNNs resolve this by reducing the number of free parameters and allowing the network to be deeper by convolutions.
    \item Pooling Layers: Pooling layers reduce the dimensions of the data by combining the outputs of clusters from the previous layer into a single node in the next layer. Popular pooling options include max pooling and average pooling, that compute the maximum value and average value of the clusters at the prior layer, respectively. The benefits of pooling are to reduce computational costs by reducing the number of free parameters as well as alleviate over-fitting by generalizing the input clusters for the following layers.
    \item Fully Connected Layers: Fully connected layers connect the nodes from the previous layer to the nodes specified for the next layer. This is the final step to generalize the outputs from convolutional and pooling layers and provide outputs for image classification tasks.
\end{itemize}

The advantages of applying Convolutional Neural Networks to image classification are (1) requires less prior processing work e.g. feature extraction, (2) reduces dimensional complexity and computational cost, (3) mitigates the over-fitting problem and (4) provides human-level correctness.

\subsubsection{AlexNet}
AlexNet \cite{balex} is considered one of the most influential architectures in computer vision after achieving nearly 50\% error rate reduction in the ImageNet challenge, having spurred many more papers published employing CNNs and GPUs to accelerate deep learning.

The main improvements of AlexNet are implementing Rectified Linear Units (ReLUs) and Dropout Layers in the network architecture.

\begin{itemize}
    \item ReLU Layers: After convolution operations done by convolutional layers, it is convention to apply a nonlinear layer (activation layer) to introduce non-linearity to the model. Since the convolutions consist of linear operations like multiplications and summations, it is important to make the model nonlinear for complex image classification tasks. For the traditional nonlinear operations including tanh and sigmond, AlexNet applies ReLUs ($f(x) = max(0, x)$) which costs much less computational time and alleviates the vanishing gradient problem without compromising much accuracy. 
    \item Dropout Layers: In Pooling Layers, over-fitting in training process occurs when the parameters (weights) are tuned too much to over-fit the samples, resulting in a poorly performed model on new samples. The idea of dropout is to randomly set a layer of activations to be 0. Dropout layers further alleviate the issue of over-fitting by assuming that a well performed model should provide good classifications even if some random activations are dropped out.
\end{itemize}

\subsubsection{ResNet}

ResNet \cite{bres} resolves the issue that deep networks suffer from that saturate and degrade accuracy while increase the number of layers by using skip connections that are also known as residuals to identity blocks which form basic blocks in its structure along with convolutional blocks.

\begin{figure}[htbp]
\centerline{\includegraphics[scale=0.4]{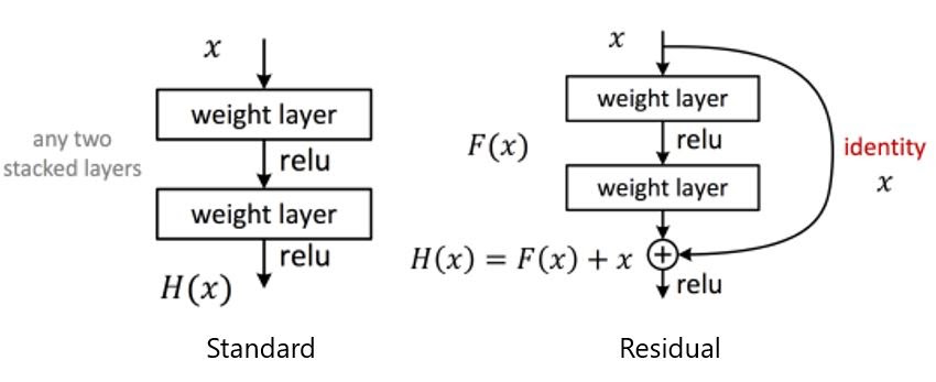}}
\caption{Comparison of a standard block and a residual block.}
\label{fig:resblock}
\end{figure}

As shown in Fig.\ref{fig:resblock}, residual blocks add a connection between network layers and the features from the previous layers. Skip connections allow the features to be easily propagated through the network. The summation of the features from previous layers increases the accuracy of the network.

\subsubsection{DenseNet}

Similar to ResNets, DenseNets \cite{bdense} also use shortcut connections in the network structure. DenseNets extend the idea of skip connections to every layer and provide a much more densely connected architecture.

The main fundamental difference is that DenseNets use concatenated feature maps from all preceding layers rather than summation of the previous layers in ResNets.

The advantages of DenseNet include (1) uses fewer parameters for training and (2) reduces computational cost. For instance, a ResNet with 101 layers can achieve a similar accuracy with a DenseNet with 201 layers. However, Densenet has only 45 \% of the number of the parameters used in ResNet and can be trained nearly twice as faster than ResNet.

\subsubsection{MobileNet}

Mobile devices are a massive market for deep learning models. Due to the trade off between the number of layers in terms of accuracy and the memory cost, MobileNets \cite{bmobv1} \cite{bmob} have become popular for deployment on hardware.
 
The main idea of MobileNets is to use depth-wise separable convolutions instead of point-wise convolutions like in other CNN models, represented in Fig.\ref{fig:mob}.
 
\begin{figure}[htbp]
\centerline{\includegraphics[scale=0.4]{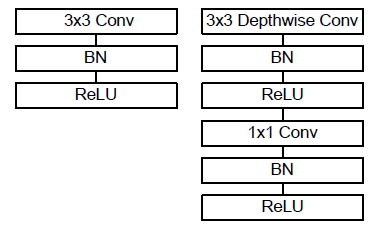}}
\caption{Comparison of a standard convolution and a depthwise separable convolution in MobileNets.}
\label{fig:mob}
\end{figure}

MobileNets apply Batch Normalization (BN) and ReLUs after each convolution. When the kernel size of the convolution operation is $3\times3$, nearly 9 times less computation power will be achieved.

\section{Results and Discussion}

In this section, we systematically evaluate the performance of each prior trained models in PyTorch framework \cite{bpytorch} which are introduced in the previous section. The models we assess include a regular 4-layer CNN model, ResNet models with 34, 50 and 101 layers, respectively, a DenseNet model with 161 hidden layers, an AlexNet model and a MobileNetV2 model. 

A data set containing 2000 samples is used for the training and testing processes for each model. The samples in the data set are randomly selected images with human-generated labels from LADI data set where half of the samples are labeled as ``damage: flood/water", and the other half are labeled as other kinds of damages or no damage. The goal of training and testing different models with such a data set is to provide a binary classifier to classify whether an image contains flooding or not. Note: The data set is not a fixed database for all models. Each time before training a model, we randomly select 2000 samples with a fixed ``flooding : non-flooding" (50\% : 50\%) ratio from the LADI data set. In this way, we can mitigate the bias of over-fitting and under-fitting by feeding stochastic-ally chosen samples into models for our experiments each time.

We split our data set with 80\% of samples (1600 images) for training and 20\% of samples (400 images) for testing. We train each model for 30 epochs and test it accordingly.

In the testing process, we first get the machine generated labels by the model based on the predictions of our binary classifier. They are then compared to the ground truths generated by human beings in the LADI data set. The binary classifier returns label 1 for flooding images and 0 for non-flooding images. If the predicted labels match the ground truths, the detection of a flooding or a non-flooding image is successful. In this way, we get accuracy scores for different classifiers.

Table~\ref{tab:acc} compares the accuracy and size of the models trained on 30 epochs for the randomly generated data set as binary classifiers for flood detection in images. The regular 4-layer CNN model performs the worst and gets the largest size in the 7 prior trained models. ResNet models achieve good accuracy and occupy relatively small memory space. As the number of layers increase, the accuracy and size of the model also increases. ResNet 101 model achieved the best accuracy of 79\% among all the models we have trained. DenseNet with 161 hidden layers obtained a satisfactory accuracy of 76\% as well as maintain a relatively small size. AlexNet, considered as one of the most influential models in Computer Vision, also gets 76\% accuracy but with a huge size. In contrast, MobileNet V2, although does not get an outstanding accuracy score, it has the smallest size of merely 17 megabytes, illustrating its potential to be deployed on a hardware, such as mobile devices, embedding systems and web servers.

\begin{table}[htbp]
\caption{Accuracy (\%) and Size (MB) of 4-Layer CNN, ResNet34, ResNet50, ResNet101, AlexNet, DenseNet and MobileNetV2 Models Trained for 30 Epochs}
\begin{center}
\begin{tabular}{ccc}
\hline
\textbf{Model} & \textbf{Accuracy (\%)} & \textbf{Size (MB)} \\ \hline
4-Layer CNN    & 68                     & 3794               \\ \hline
ResNet34       & 72                     & 163                \\ \hline
ResNet50       & 75                     & 180                \\ \hline
ResNet101      & 79                     & 325                \\ \hline
AlexNet        & 76                     & 539                \\ \hline
DenseNet       & 76                     & 203                \\ \hline
MobileNetV2    & 73                     & 17                 \\ \hline
\end{tabular}
\label{tab:acc}
\end{center}
\end{table}

Our next experiment is to provide predicted results by the trained model on our test samples. Fig.~\ref{fig:pred} presents the predictions of ResNet 101 model, which obtains the best accuracy as shown in Table~\ref{tab:acc}, on 15 random images in the test set. 

\begin{figure}[htbp]
\centerline{\includegraphics[scale=0.4]{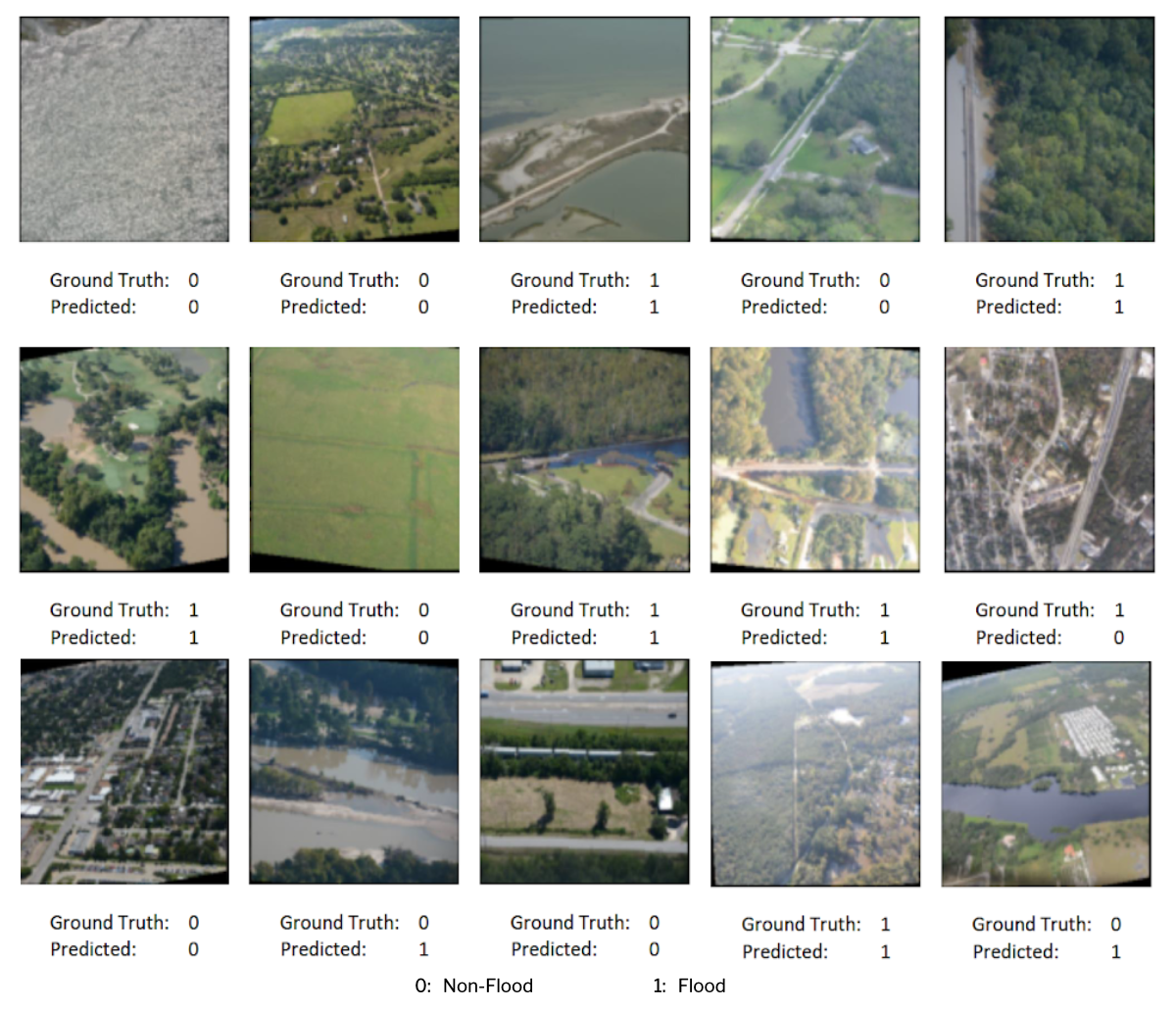}}
\caption{Predictions of ResNet 101 Model on 15 test images.}
\label{fig:pred}
\end{figure}

Among these 15 test images, only 3 images are falsely classified: one in the second row, last column; another in the last row, second column and the last one in the last row, last column. If we look at the image in the second row and the last column, we can find that it includes highways, roads, buildings and lands. Although it is labeled as flooding, the flooding pattern is too subtle to be discovered, or it is incorrectly labeled by human beings. Similarly, the second false classified image, in the second row, last column, could be falsely labeled. We can see that the water invades the boundary of the land, but the image is labeled as non-flooding. In this case, the classifier can also serve as a filter to find out suspiciously labeled images and promote further data cleaning and enhancement of LADI data set. The last wrongly classified image in the last row, last column demonstrates the limit of our current ResNet 101 classifier which requires further training and improvement.

Accuracy is a good metric to measure the proportion of correctly classified instances over all the samples in the test set. However, to evaluate a classifier, accuracy is not always the pivotal score. In some cases, a classifier can get a good accuracy but not a good performance in real world problems. Suppose a classifier always predicts 0 for a binary classification task with a test set containing 90\% of samples labeled as 0 and 10\% as 1. The accuracy is high, but this classifier will not perform well. To eliminate the deficiency of accuracy, below is a confusion matrix of our ResNet 101 model with counts and ratios for True Positives (TP), False Positives (FP), True Negatives (TN) and False Negatives (FN) in Fig.~\ref{fig:cm}.

\begin{figure}[htbp]
\centerline{\includegraphics[scale=0.2]{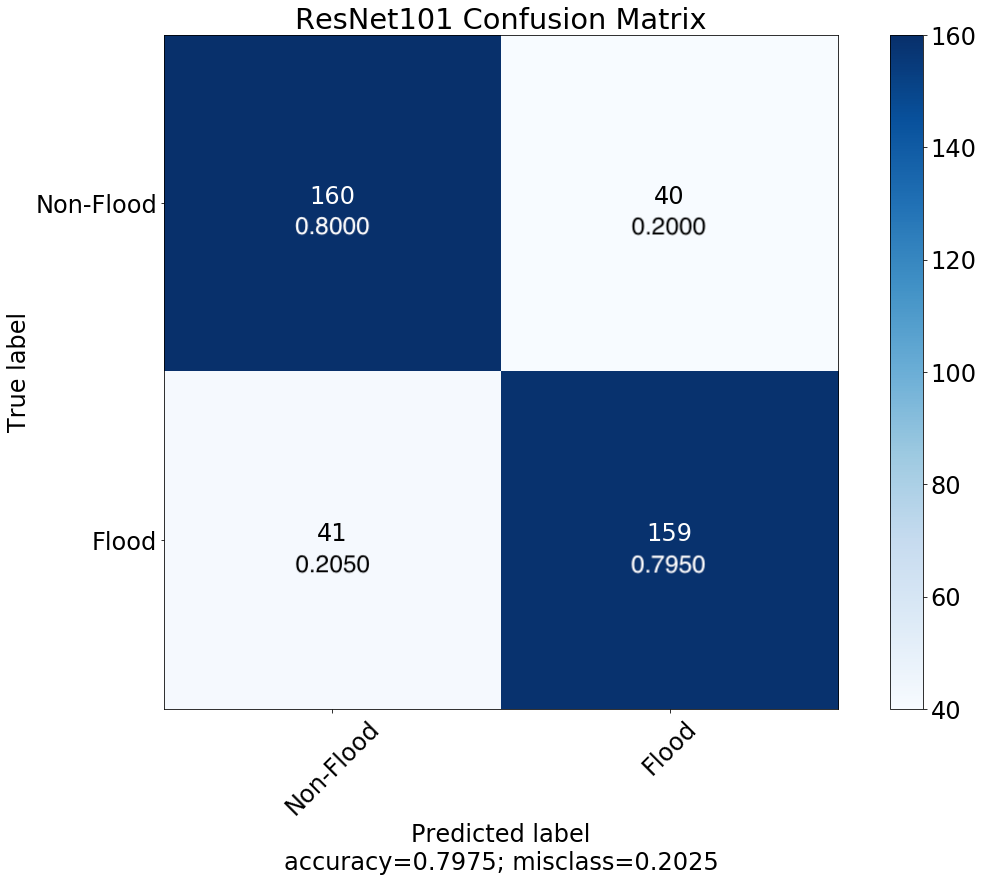}}
\caption{Confusion matrix of ResNet 101 model.}
\label{fig:cm}
\end{figure}

From Fig.~\ref{fig:cm}, we can see the 4 outcomes of a binary classification: 
\begin{itemize}
\item True Positives: data instances labeled as positive (flooding) that are actually positive (flooding).
\item False Positives: data instances labeled as positive (flooding) that are actually negative (non-flooding).
\item True Negatives: data instances labeled as negative (non-flooding) that are actually negative (non-flooding).
\item False Negatives: data instances labeled as negative (non-flooding) that are actually positive (flooding).
\end{itemize}

Based on the four outcomes in the confusion matrix, we can use precision and recall metrics to evaluate the model.
\begin{itemize}
\item Precision: ability of a classification model to return only relevant instances.
\item Recall: ability of a classification model to identify all relevant instances.
\end{itemize}

The equations for precision and recall are shown below:
\begin{center}
    $Precision=\frac{TP}{TP+FP}$,    $Recall=\frac{TP}{TP+FN}$
\end{center}

In our binary classification, precision is the ratio of the flooding samples correctly identified over the sum of the flooding samples correctly identified and the instances incorrectly identified as flooding. If the precision is high, the images that the classification model classified as positives are more likely to be actually positives. Recall is the ratio of the flooding samples correctly identified over the sum of the flooding samples correctly identified and the flooding samples incorrectly identified as non-flooding. If the recall is high, the classification model is more likely to capture all flooding images in the data set and label them as flooding.

The precision and recall of our ResNet 101 model are 79.5\% and 79.9\%, respectively. The high precision and recall scores indicate our ResNet 101 model is a capable and precise flooding imagery classifier.

In this section, we demonstrate our experimental results of our models and discuss several interesting outputs. The results of our models are considered to be exceptional, but we expect further improvements of the classifiers as well as the LADI data set. The next section will give a summary of our project and offer a prospect of the future work.

\section{Conclusion and Future Work}
\subsection{Conclusion}
The LADI project is designed to develop a useful and efficient tool to quickly respond to a disaster based on imagery classification and detection. The model  we developed would become a part of the tool to detect and classify images in the LADI data set . Given LADI data set, our model processes the input images and classifies them if they include flooding or not. The result could be used for further disaster responders.

In this paper, we implemented a binary classifier for flooding imagery classification based on the LADI data set. We successfully trained various convolutional neural networks including a regular CNN model, an AlexNet, ResNets, a DenseNet and a MobileNet with satisfactory accuracy scores.

From our experimental results, we obtained a ResNet 101 model with the highest accuracy of 79\% as well as exceptional precision and recall scores of nearly 80\%, indicating the good performance of our CNN models in the disaster imagery classification tasks. We achieved a MobileNetV2 model which takes only 17 megabytes, illustrating the potential of MobileNets for deployment on hardware devices.

By comparing human generated labels as the ground truths and the model predicted labels, we obtained the accuracy scores of various classifiers we have trained. By examining the True Positives (TP), False Positives (FP), True Negatives (TN) and False Negatives (FN) in our classification results, we inspect our binary classifier more deeply. The outstanding precision and recall scores of our classifier indicate the capability and precision of our binary classifier.

\subsection{Future Work}
Our deep learning models are available publicly to enable potential end-users to adopt, modify, and improve our already existing models. Since we are the first team to develop classifiers for this flooding classification set, our code and documentation will be used in the future for a class taught by MIT.

In the future, focusing on improving the accuracy of MobileNet for later hardware deployment would be beneficial. MobileNetV2 achieved an accuracy of 73\% with a size of only 17MB. In comparison, our next highest size is for ResNet34 with 163MB. This is an extremely large gap, making MobileNet the most suitable for deployment on embedded hardware.  

Other future work may also include extending our binary classifier to multi-classifier and multi-label classifier. Furthermore, because there have been images that are falsely classified by humans, our trained classifiers may also aid in finding suspicious human generated labels. Our classifier can essentially help filter out the images with mismatched labels for future data cleaning. 

Future iterations should place emphasis on the system deployment to embedded hardware. Using commercially embedded development platforms such as Raspberry Pi, Intel Neural Compute, or Google Edge TPU is highly recommended when developing the device that deploys the trained deep learning models. The embedded hardware should be able to perform online detection and classification; therefore, enabling institutions such as FEMA to assess damage prior to arriving on-site. Drones or weather balloons are recommended in order to retrieve aerial views/images of the specified area.

\section*{Acknowledgment}

We would like to thank Dr. Jeffrey Liu and Andrew Weinert for sponsoring this project, as well as providing the LADI dataset, weekly discussions, and guidance with working on the Deep Learning Models. We would also like to thank Dr. Marc Rigas for forming this team, providing weekly guidance, and looking over the direction of our work.

\end{document}